\pdfoutput=1
\documentclass[11pt]{article}

\usepackage[final]{acl}

\usepackage{times}
\usepackage{latexsym}

\usepackage[T1]{fontenc}
\usepackage[utf8]{inputenc}

\usepackage{microtype}

\usepackage{inconsolata}
\usepackage{url}

\usepackage{graphicx}
\usepackage{booktabs}

\title{Interpreting Bias in Large Language Models: A Feature-Based Approach}

\author{Nirmalendu Prakash \\
  Singapore University of \\Technology and Design \\
  \texttt{nirmalendu\_prakash@mymail.sutd.edu.sg}  \\\And
  Roy Ka-Wei Lee \\
  Singapore University of \\Technology and Design  \\
  \texttt{roy\_lee@sutd.edu.sg}\\}

\begin{document}
\maketitle
\begin{abstract}
% Few lines about biases, how they have been tackled in earlier works. How this approach differs. Why do we think this is better.
Large Language Models (LLMs) such as Mistral and LLaMA have showcased remarkable performance across various natural language processing (NLP) tasks. Despite their success, these models inherit social biases from the diverse datasets on which they are trained. This paper investigates the propagation of biases within LLMs through a novel feature-based analytical approach. Drawing inspiration from causal mediation analysis, we hypothesize the evolution of bias-related features and validate them using interpretability techniques like activation and attribution patching. Our contributions are threefold: (1) We introduce and empirically validate a feature-based method for bias analysis in LLMs, applied to LLaMA-2-7B, LLaMA-3-8B, and Mistral-7B-v0.3 with templates from a professions dataset. (2) We extend our method to another form of gender bias, demonstrating its generalizability. (3) We differentiate the roles of MLPs and attention heads in bias propagation and implement targeted debiasing using a counterfactual dataset. Our findings reveal the complex nature of bias in LLMs and emphasize the necessity for tailored debiasing strategies, offering a deeper understanding of bias mechanisms and pathways for effective mitigation.
%Large language models, such as GPT, are known to exhibit social biases, including gender bias. While recent studies have proposed debiasing techniques using causal analysis, we demonstrate that the components identified through these methods might have a significant effect but may not necessarily be the actual "source" of bias. We propose an approach of thinking in terms of features in order to trace bias from origin to output, utilizing techniques from mechanistic interpretability research. In contrast to the previous approaches, this is a case-by-case approach to bias studies. Our study investigates gender bias in three recent open-source LLMs: LLaMA3 8B, LLaMA2 7B, and Mistral 7B v0.3. We use the professions dataset and through our experiments, demonstrate that these models exhibit common behaviour. 
\end{abstract}

\section{Introduction}
% About social biases.

% Debiasing approaches. Our fundamental improvement on that.

% Interpretability techniques. How have we used them.

% Contributions: 1) Meahanistic approach to understanding biases. 2) Experiments to proove/disprove hypothesis. 3) Generalization of understanding to different family of models.
\begin{figure*} 
    \centering
    \includegraphics[scale = 0.73]{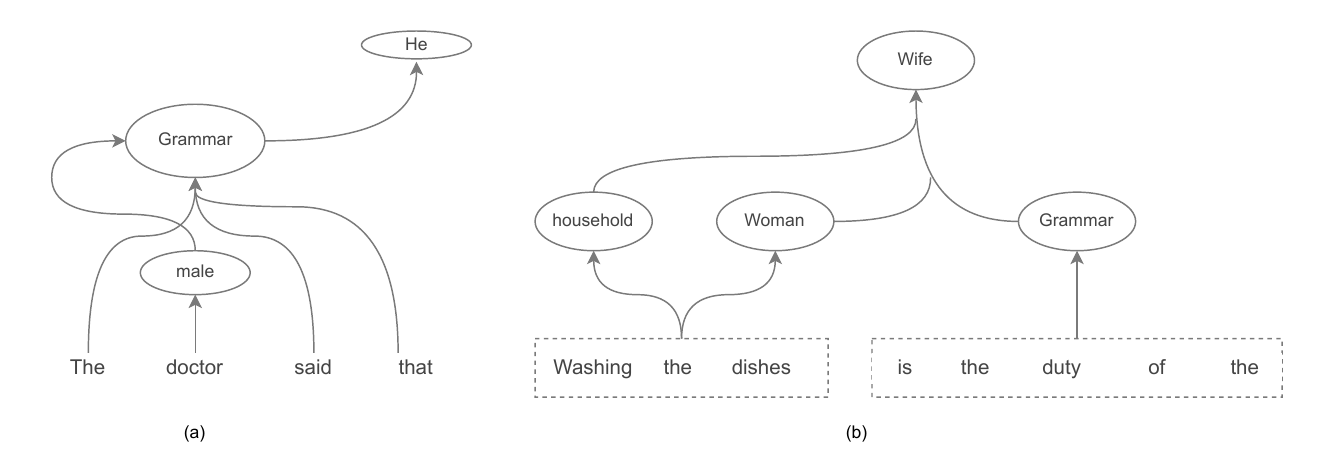}
    \caption{Conceptualized feature map in LLMs for prompts: a) \textit{"The doctor said that"} and b) \textit{"Washing the dishes is the duty of the"}. }
	\label{fig:mental model}
\end{figure*}

%Large Language Models (LLMs) have been shown to excel in numerous natural language processing (NLP) tasks such as text generation, question answering and translation. Models like OPT \cite{zhao2018gender} and LLaMA \cite{touvron2023llama} are trained on extensive and varied datasets that include webpages, Wikipedia, books, scientific papers, and other online content. This wide array of data ensures a comprehensive representation of global knowledge. However, it also presents a significant challenge: while it captures a broad and diverse range of ideas, it simultaneously exposes the models to inherent social biases.
\textbf{Motivation.} Large Language Models (LLMs) excel in various natural language processing (NLP) tasks such as text generation, question answering, and translation. Models like Mistral \cite{jiang2023mistral} and LLaMA \cite{touvron2023llama}, trained on diverse datasets including webpages, Wikipedia, books, and scientific papers, provide a comprehensive representation of global knowledge. However, this diversity also exposes these large models to inherent social biases.

%In recent years, the NLP community has prioritized studying biases in LLMs. Early work by \cite{bolukbasi2016man} revealed gender and ethnic
%biases in word embeddings like Word2Vec and GloVe. This trend of identifying biases continued with more complex models like BERT, where researchers examined how biases are encoded and propagated \cite{kurita2019measuring,may2019measuring}.
Addressing biases in LLMs has become a priority for the NLP community. Early work has revealed gender and ethnic biases in word embeddings like Word2Vec and GloVe \cite{bolukbasi2016man}, and subsequent research extended this to more complex models like BERT \cite{kurita2019measuring,may2019measuring}. \citet{vig2020investigating} introduced the use of causal mediation analysis to identify bias-causing neurons or attention heads, demonstrating that in decoder-only models such as GPT-2, gender bias effects are sparse. More recently, \citet{chintam2023identifying} expanded this approach to include fine-grained causal mediation called ACDC \cite{conmy2023automated} and a combinatorial approach, DiffMask \cite{de2021sparse}, for bias analysis. These studies focus on identifying components most effective in reducing bias, measured using a counterfactual dataset through causal mediation. Nevertheless, these studies do component identification in a generalised fashion. As components that contribute to one form of gender bias may not contribute to another, we propose a case-by-case study.

Various studies have have explored debiasing techniques for masked language models such as BERT \cite{meade_2022_empirical} and larger models like LLaMA and OPT \cite{ranaldi2023trip}, using methods like LoRA training on counterfactual datasets. Layer-wise bias benchmarking of original and debiased models has also been conducted \cite{prakash2023layered}. Nevertheless, these studies have applied debiasing in a general fashion by training adapter heads. We propose targeted approach to debiasing and demonstrate the efficacy by applying counterfactual finetuning on identified components only.

\textbf{Research Objectives.} In this research, we aim to address existing gaps by investigating the roles of neural components in producing biased outputs. For instance, we question whether an attention head introduces bias or merely replicates it. Inspired by a recent study that applied a feature evolution model to identify a ``refusal feature'' for harmful requests in LLMs \cite{refusalstudy},in the context of our study, we define a features as abstract concepts related to bias. For example, when examining gender bias, a feature might represent a latent concept such as "male" or ``female'' related to gender bias. We call the supposed evolution of such features within the model as ``feature map''. 

Figure \ref{fig:mental model} illustrates a feature map in LLMs for two prompts, where we have mapped the input to output with intermediate features in between. These features may or may not align with the actual processing of LLMs, necessitating iterative refinement of our map. For example, in the prompt \textit{``The doctor said that''}, for a biased output ``he'', we hypothesize that an intermediate feature ``male'' is derived from ``doctor''. This feature, combined with a grammatical feature representing sentence structure, produces the final output ``he''. Similarly, for the sentence \textit{``Washing the dishes is the duty of the''}, we hypothesize that \textit{``Washing the dishes''} produces features such as ``household'' and ``woman''. These combine with the sentence's grammatical structure to produce the abstract feature ``wife''. This feature map guides our experiments on interpreting bias in LLMs. 

We apply our proposed approach to understand bias propagation in state-of-the-art open-source LLMs such as LLaMA-2-7B, LLaMA-3-8B, and Mistral-7B-v0.3. Interpretability techniques like activation and attribution patching are used to validate our feature-based understanding. Finally, we analyze the effects of debiasing on these LLMs.

\textbf{Contributions.} We make the following research contributions: (i) We introduce a feature-based approach to studying bias in LLMs. We form hypotheses and perform experiments to validate them on thre state-of-the-art closed-source LLMs: LLaMA-2-7B, LLaMA-3-8B, and Mistral-7B-v0.3, using templates from a professions dataset. (ii) We apply this approach to another form of gender bias outside the original dataset, demonstrating the approach's extensibility. (iii) We delineate the distinct roles of MLPs and attention heads in generating gender bias and perform targeted debiasing using a counterfactual dataset.

\section{Related Work}

% 1) LLM debiasing
% 2) Mechianistic Interpretability

\subsection{Bias Studies in NLP}
Early bias analysis in NLP focused on linear relations in word embeddings, which was later extended to analyzing sentence embeddings in encoder models such as BERT. \cite{vig2020investigating} studied the decoder-only autoregressive model GPT-2 and introduced causal mediation techniques to localize bias, examining the role of attention heads. Similarly, \cite{chintam2023identifying} investigated where gender bias is introduced in LLMs, focusing on attention heads. In contrast, our work does not limit the analysis to MLPs or attention heads. We approach the problem by first forming hypotheses based on a conceptual model of feature evolution and then performing experiments to verify these hypotheses. Our experiments are conducted on larger models such as LLaMA and Mistral, as we believe these LLMs are more likely to replicate the feature evolution model.

Several studies have also concentrated on curating datasets and metrics for bias measurement. For example, \cite{sheng2019woman}, \cite{dhamala2021bold}, \cite{kirk2021bias}, and \cite{li2020unqovering} employ open-ended text generation with prompts such as ``\textit{The woman works as}'' and measure bias using specially trained classifiers or off-the-shelf tools. Conversely, benchmarks like StereoSet \cite{nadeem2021stereoset} measures bias by calculating the probability of generating specific words or sequences. In our work, we measure the fraction of samples where the LLM prefers anti-stereotypical sentences over their stereotypical variants.

%Early bias analysis in NLP revealed linear relations in word embeddings. This was extended to analysing sentence embedding for encoder models such as BERT. \cite{vig2020investigating} study decoder only autoregressive model, GPT2 and introduce causal mediation technique to localise bias. The authors examine the role of attention heads. \cite{chintam2023identifying} also look for where in a LLM is gender bias introduced, but limit their focus on attention heads. Instead in this work, we do not limit to MLPs or attention heads and approach the problem by first forming a hypothesis based on a mental picture of features and then perform experiments to verify that. Also, we perform our experiments on larger models such as LLaMA and Mistral. We belive that LLMs are more likely to replicate the feature image. 

%On the other hand, several works have focused curating datasets and metrics for bias measurement. Some of the studies, such as those by \cite{sheng2019woman,dhamala2021bold,kirk2021bias,li2020unqovering}, employ open-ended text generation. They use prompts, such as "The woman works as" and measure bias via a specially trained classifier or by using off-the-shelf tools. Conversely, benchmarks like StereoSet gauge bias by calculating the probability of generating specific words or sequences. In this work, we measure of fraction of samples where LLM prefers anti-stereotypical sentence over the stereotypical variant.

\subsection{LLM Interpretability}
\citet{vig2020investigating} found that only a small number of attention heads in GPT-2 have a high indirect effect on bias, exhibiting a high degree of sparsity. The authors utilized causal intervention techniques, which may vary depending on the ablation type. For instance, activations can be zeroed out \cite{geva2023dissecting}, replaced with a baseline activation \cite{bau2018identifying}, or more commonly, replaced with activations from another example. This technique, increasingly referred to as activation patching \cite{ferrando2024primer}, has emerged as a popular tool for mechanistically understanding LLMs. A counterfactual dataset is preferred for patching as it does not affect the distribution and inverts the prediction. In our work, stereotypical professions are used to create counterfactual samples.

The causal link to output can be verified by studying changes in predictions using metrics such as KL divergence or logit/probability difference. Essentially, the forward pass of model computation is viewed as a directed acyclic graph, where nodes can be neurons, attention heads, MLP outputs, etc., based on the desired granularity. Interventions can be made at the nodes or edges, using a technique popularly referred to as the \textit{transformer circuits} framework \cite{transformer-circuits}. In this work,  we apply the logit difference to study MLPs and attention heads.

Recently, several works \cite{wang2022interpretability} have used edge patching to identify a ``\textit{circuit}'', which is a path in the computation graph contributing to the output. Researchers have studied circuits for common tasks such as Indirect Object Identification (IOI) \cite{wang2022interpretability}, Multiple Choice Question Answering (MMLU) \cite{lieberum2023does}, and the greater-than task \cite{hanna2024does}. These techniques have largely been applied to relatively smaller models such as GPT-2, due to concerns about computational inefficiency. To overcome this limitation, approximation techniques such as attribution patching \cite{attribpatching,kramar2024atp} have been proposed. We employ attribution patching in combination with activation patching in a targeted manner.

\section{Experimental Setup}
%\roy{[Roy: It will be better if we can give an overview of how the experiment is conducted and the intuion for doing it this way (i.g., what we hope to achieve)]}
%Using the feature map we have as basis, we form hypotheses, first about the origin and then about propagation of bias. We conduct experiments in order to verify each, and answer any new questions that arise. The dataset, LLMs and techniques required for the experiments are briefly explained here:

To explore the origins and propagation of bias in LLMs, we form and test hypotheses based on our feature map. Our experimental setup, including datasets, models, and techniques, is detailed below:

%\textbf{Datasets}: We follow the approach of \cite{chintam2023identifying}, where professions dataset \cite{NEURIPS2020_92650b2e} is used to identify bias causing model components and benchmark debiased models on professions, CrowSPairs \cite{nangia2020crows} and WinoBias \cite{zhao2018gender} datasets. As our focus is gender bias, we ignore the other forms of bias in CrowSPairs. We suppliment professions dataset with additional stereotypical professions from \cite{mattern2022understanding}. List of professions used is available in \ref{} 

\textbf{Datasets}. Following \cite{chintam2023identifying}, we use the professions dataset \cite{NEURIPS2020_92650b2e} to identify bias-causing components and benchmark debiased models on professions, CrowSPairs \cite{nangia2020crows}, and WinoBias \cite{zhao2018gender} datasets. We use an updated version of CrowSPairs \cite{neveol2022french} and focusing on gender bias, we exclude other biases in this dataset. Additionally, we incorporate more stereotypical professions from \cite{mattern2022understanding}. The complete list of professions used is provided in Appendix A4.

For debiasing, we utilize the Pandas \cite{qian2022perturbation} dataset, which perturbs natural sentences by swapping words. For example, \textit{He had no passion about anything} becomes \textit{She had no passion about anything}. We filter this dataset to include only binary gender-related words (listed in Appendix A5), resulting in 18,380 samples.

%In all banchmarks, bias is measured as the ratio of samples where model prefers anti-stereotypical sentence over the stereotypical variant. 

\textbf{Measuring of Bias}. Bias is measured by the ratio of samples where the model prefers anti-stereotypical sentences over stereotypical ones.

%To debias the models, we use Pandas \cite{qian2022perturbation} dataset. This dataset is developed by perturbing natural sentences on a target attribute by swapping a word. For example, for a sentence, \textit{He had no passion about anything.} is perturbed to create \textit{She had no passion about anything.}, where the target attribute is woman and the swapped word is \textit{he}. We filter Pandas dataset with a set of target words related to binary genders (listed here \ref{}), resulting in 18,380 samples.

%\noindent \textbf{LLMs}: We perform our study on some of the latest open source decoder only language models: LLaMA-2-7B \cite{touvron2023llama} and LLaMA-3-8B \cite{llama3modelcard} and Mistral 7B v0.3 \cite{jiang2023mistral}.

\textbf{LLMs}. Our study employs state-of-the-art open-source decoder-only LLMss: LLaMA-2-7B \cite{touvron2023llama}, LLaMA-3-8B \cite{llama3modelcard}, and Mistral-7B-v0.3 \cite{jiang2023mistral}.

% \noindent \textbf{Information Flow} We use the importance metric porposed by \cite{} to identify important attention heads which contribute to bias. This works much faster for LLMs used compared to ACDC \cite{}.

% 4) libraries - transformerlens (issue- differs in 2nd decimal compared to hf), distance metric for information flow

%\noindent \textbf{Activation Patching}: 
% Activation patching or Causal Mediation Analysis was originally proposed by \cite{NEURIPS2020_92650b2e} and has used in the mechanistic interpretability work \cite{wang2022interpretability,lieberum2023does,stolfo2023mechanistic}. It is a technique of swapping specific internal activations of a model typically with activations from a counterfactual sample. A metric such as logit difference is used to measure the effect of this perturbation and attributed to the components patched. The components can be a neuron, MLP output, Attention head output or even the residual stream. We use this technique in a targetted manner to verify hypothesis, as doing activation patching on all components of LLMs is compute inefficient. Henceforth, we drop the term activation and use only "patching".
%We perform activation patching on a set of MLP outputs. Henceforth, we drop the term activation and use \textit{patching} to refer to this technique. 

\textbf{Activation Patching}. We conduct activation patching on MLP outputs. For brevity, we refer to this technique simply as \textit{patching}.

%\noindent \textbf{Attribution Patching}: 
%Attribution patching uses gradient based approximation to activation patching, requiring two forward passes and oen backward pass through the model. We use attribution patching to determine important attention heads, followed by activation patching using the insights from attribution scores.

\textbf{Attribution Patching}. Attribution patching, a gradient-based approximation of activation patching, involves two forward passes and one backward pass. We use this method to identify important attention heads, which are then targeted with activation patching based on attribution scores.

%\noindent \textbf{Logit order}: We are interested in a set of model components which when patched with an anti-stereotypical sample cause the model to assign higher probability to the anti-stereotypical pronoun. For this, we form hypothesis regarding the working of the model, apply patching to a set of components and observe any change in pronoun order. We use the logit values instead of probability and refer to the stereotypical order of he and she as ``logit order'' and reversal of this order as ``logit reversal''.

\textbf{Logit Order}. We investigate model components that, when patched with anti-stereotypical samples, increase the probability of anti-stereotypical pronouns. Hypotheses regarding model behavior are tested by applying patching and observing changes in pronoun order. We use logit values to define the stereotypical order of ``he'' and ``she'' as ``logit order'', with reversals termed ``logit reversal''.

% 5) Activation Patching
% 6) Information Flow
% 7) what do we mean by pronoun order
% 8) Finetuning dataset

\section{Bias Analysis}

% Inspired by SEAT, "The doctor said that" should be closer to "The man said that". "I met a doctor yesterday" closer to "I met a man yesterday". Similarly, "The nurse said that" is closer to "The woman said that", at the last position of the decoder model. Though SEAT applies to encoders such as BERT. 
% \textbf{Context} : \cite{} demonstrate that refusal in LLMs is mediated by a single direction and ablating representation in this direction hinders the models' ability to refuse harmful requests. The authors hypothesize that there is a single intermediate feature that causes refusal across a wide range of harmful prompts. A higher level feature such as refusal is thought to evolve from intermediate features such as dangerous, illegal, unethical etc.
% \textbf{Hypothesis} : Our hypothesis is that there is a gender feature which is mediated by There is a "male" and "female" direction in the embedding space and "male" professions are closer to "male" direction and "female" professions closer to "female" direction. We believe that gender is not an abstract feature(maybe true for a sentence such as "the doctor said that" but is it true for "the monument was designed by", where we check for him/her. This could take the path design->designer->male->him, and require a separate study) and should be well established in the earlier layers by the model.

% Take he/she, man/woman  in the same templates, these could roughly be directions. Remove projections of male,female profession embeddings in early layers alogn these directions, does it work?

\begin{figure}[ht!] 
    \centering
    \includegraphics[scale = 0.75]{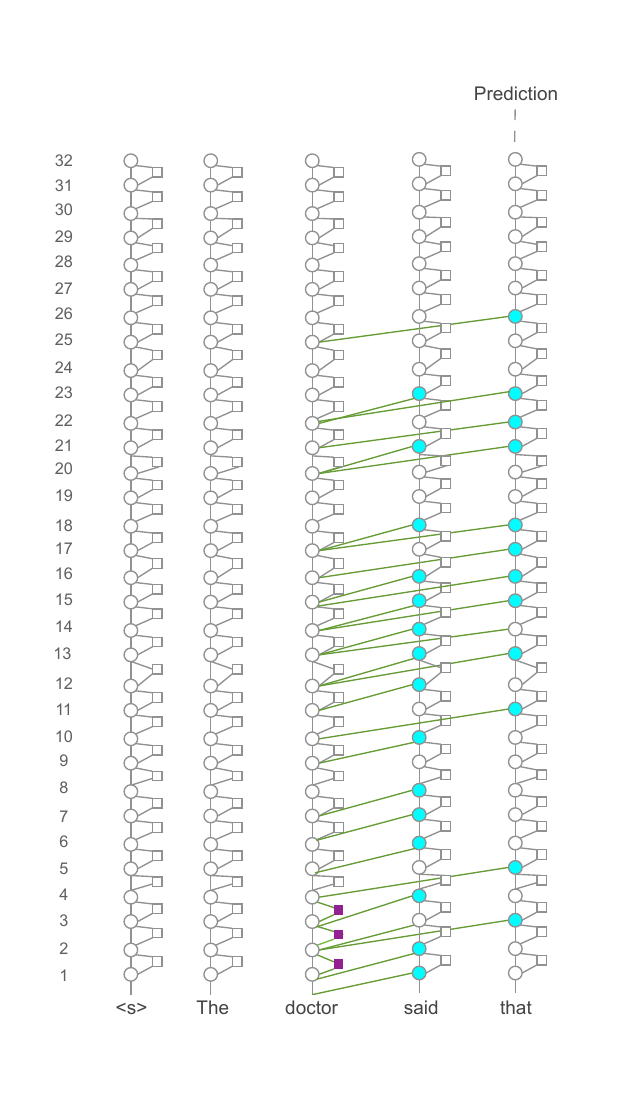}
    \caption{Bias Propagation for the prompt ``\textit{The doctor said that}'' on LLaMA-3-8B. Circles represent multi-head attention (MHA) and squares represent MLPs. Purple squares represent MLPs associated with the feature. Blue circles represent MHA responsible for copying and green lines show the corresponding connections. Digits on the left represent layer indexes.}
	\label{fig:bias_propagation}
\end{figure}

\subsection{Where is Gender Bias Introduced?}

To identify the specific layers where gender bias originates in LLMs, we systematically analyze how biases develop and propagate through the network. We begin by examining partial sentences with stereotypical gender associations and observe how the model’s preferences for pronouns like ``he'' or ``she'' are formed. By patching residual streams and analyzing MLP activations and attention heads, we trace the development of bias from early to later layers, aiming to pinpoint the critical stages where bias is introduced.

\textbf{Hypothesis}: For partial sentences such as \textit{``The doctor said that''}, LLMs tend to prefer \textit{``he''} over \textit{``she''}. We hypothesize that gender bias in such contexts is derived from the early layers of the model rather than being an abstract feature formed in later layers.

%\textbf{Validation}: First, we do a preliminary check by analysing few randomly picked samples. We sample stereotypical male and female professions and templates from the dataset and then patch residual streams at early layers starting from the first layer at the profession token position. We observe that this causes logit order to reverse. Next, instead of patching the residual stream, we patch the MLP activations and attention heads seperately. We observe that MLPs at lower layers are responsible for bias.
%Encouraged by the above results, we perform the experiment on all samples in two steps:
%\begin{enumerate}
%  \item We pick a stereotypical male profession, \textit{wrestler} and a female profession, \textit{nanny} and insert these in each of the templates to create two sentences. Model is prompted with each sentence and MLP activations recorded for each layer at the profession token position. For some of the words such as nurse, we observe that it is tokenized into more than one token. We take the last token activation for such words.  
%  \item For each template and for each profession, we patch the model with corresponding anti-stereotypical activation saved in step1 and note the difference in logits.
%\end{enumerate}

\textbf{Validation}: To test this hypothesis, we pick a template, a male profession and a female profession at random and patch residual streams at early layers, starting from the first layer at the profession token position. We do this five times and observe that this intervention often reverses the logit order. Next, we separately patch MLP activations and attention heads, observing that MLPs at lower layers are primarily responsible for the effect.

Encouraged by this finding, we conduct a comprehensive experiment on all samples in two steps:
\begin{enumerate}
\item We select a stereotypical male profession (\textit{``wrestler''}) and a female profession (\textit{``nanny''}) and insert them into various templates to create sentences. The model is prompted with each sentence, and MLP activations are recorded for each layer at the profession token position. For words tokenized into multiple tokens (e.g., \textit{``nurse''}), we use the last token activation.
\item For each template and profession, we patch the model with the corresponding anti-stereotypical activation saved in step 1 and note the difference in logits.
\end{enumerate}

\begin{figure}[ht]
    \centering
    \includegraphics[scale = 0.65]{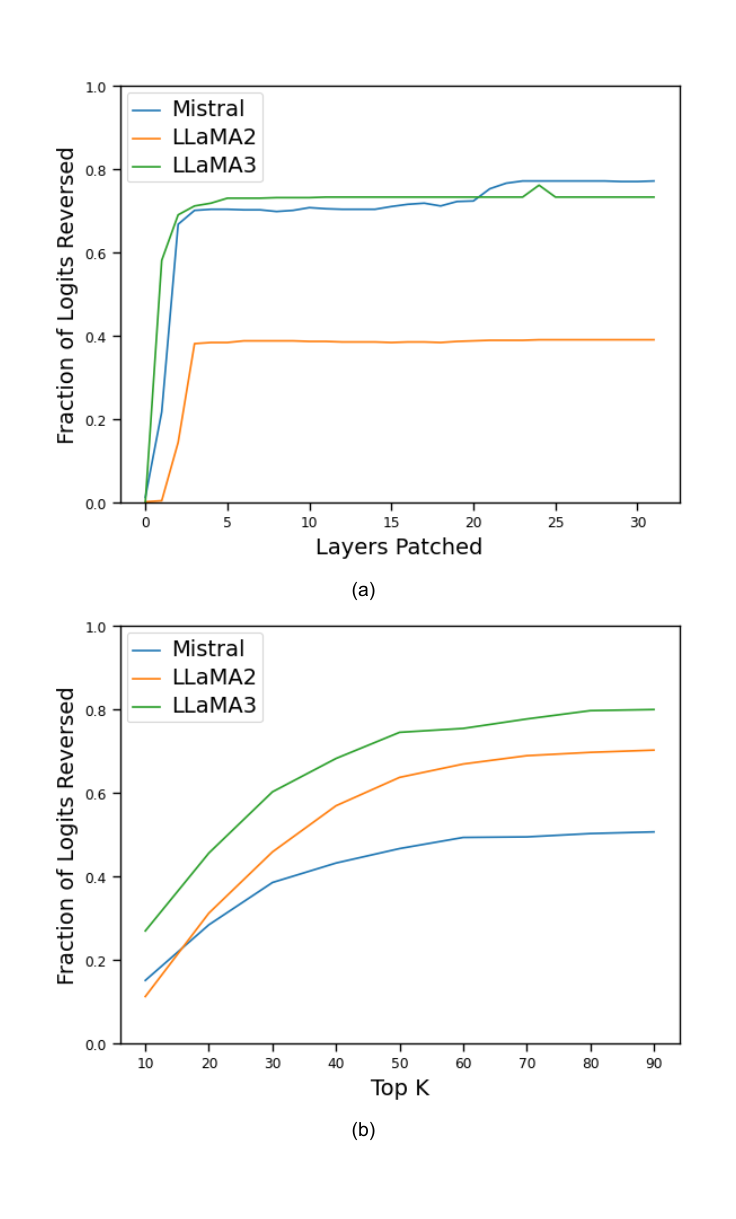}
    \caption{(a) Fraction of samples where logit is reversed vs MLP layers patched (b) Fraction of samples where logit is reversed vs top K value.}
	\label{fig:logits_reversed}
\end{figure}

\textbf{Observations}: We observe bias reduction across layers, with logit reversal occurring in many samples from layers 1 to 6 for LLaMA-3-8B and Mistral-7B. For LLaMA-2-7B, individual layer patching causes reversal in fewer samples. We then patch multiple layers incrementally, starting from the first layer MLP. Figure \ref{fig:logits_reversed}a shows the fraction of samples where logit order is reversed by patching additional layers. Patching layers 1 to 5 is sufficient to reverse logit order in 75\% of samples for LLaMA-3-8B and 73\% for Mistral-7B, with fewer reversals observed for LLaMA-2-7B.

For cases where logit reversal does not occur, we hypothesize that patching the last token position of the profession term may be insufficient, with information being copied from other locations. For example, \textit{``entrepreneur''} is tokenized as three tokens (\textit{``entrepr''}, \textit{``ene''}, and \textit{``ur''}). We tweak our experiment by finding same-length opposite-gender professions (e.g., pairing \textit{``entrepreneur''} with \textit{``librarian''}) and performing MLP output patching at all profession tokens. This adjustment results in logit order reversal.

\subsubsection{Does Patching Lead to Reversal of Profession Information?}
To verify whether patching the MLP outputs retains context about the profession, we use greedy generation, selecting the top predicted token at each step. We perform this with layers 1 to 5 MLP outputs patched at the profession token position. For cases where we observe reversal from the previous experiment, we generate 30 tokens for random 20 samples for each LLM,using greedy decoding and analyze the generated text for the presence of the profession term and the reversal of pronouns.

 In most cases, the profession term appears in the generated text, and pronouns are reversed as expected. In some cases, the association with profession term is ambiguous. These observations indicate that patching retains the profession information while mitigating bias. The results are available in Table \ref{tab:patched_mlp_gen_metrics}.

%Having identified the origin of bias, we now investigate how it propagates to the output. Table \ref{} shows sample patched generations for LLaMA-3-8B.

%Our initial setup using the last profession token continues to be effective, and we verify if patching these MLP outputs retains context about the profession.
%To verify whether the model still retains information about profession, we use greedy generation, where we take the top token predicted at each step, with layers 1 to 5 MLP outputs at the profession token position patched. We do this for sample "successful" cases from above experiment, and generate 100 tokens for each sample. We look for profession term in the generated text and reversal of pronouns. We note that in most cases, profession token appears in the generated tokens and pronouns are reversed. In the remaining cases, the generated text can be associated to the professions. After having identified the origin of bias, we investigate how this propagates to the output. Table \ref{sec:appendix} shows sample patched generations for LLaMA-3-8B.

\begin{table*}[ht!]
    \centering
    \small
    \begin{tabular}{lp{12cm}} 
       \toprule
       \textbf{Variant} & \textbf{Text} \\
       \midrule
       Prompt & \textit{The officer said that} \\
       Pretrained & \textit{the accused had been arrested in connection with the murder of a 25-year-old man ...}\\
       Patched & \textit{she was not a suspect in the true crime case. ...}\\
       \midrule
       Prompt & \textit{The architect yelled that}\\
       Pretrained & \textit{he was going to kill himself. He was standing on the roof of the building. ...}\\
       Patched & \textit{she is a thief. The architect yelled that she is a thief. ...}\\
       \midrule
       Prompt & \textit{The childcare worker slept because}\\
       Pretrained & \textit{she was tired. She was not aware that the child was in the room. ...}\\
       Patched & \textit{he was supposed to be on duty. ...}\\
       \midrule
       Prompt & \textit{The manager wished that}\\
       Pretrained & \textit{the team had a better record, but he was pleased ...}\\
       Patched & \textit{she was a little more like the one that she had hired. ...}\\
       \bottomrule 
    \end{tabular}
    \caption{Sample text generation using LLaMA-3-8B with and without patching  MLP outputs for layer 1 to 5.}
    \label{tab:mlp_patched_generation}
\end{table*}

% \subsection{Bias origin at lower MLPs at profession position --> copied by attention heads at lower and middle layers --> decipher pronoun by middle mlps at final position --> no role of upper layer mlps --> prediction?}
\subsection{How does Bias Propagate?}
%We do a preliminary analysis using logit lens technique of using unembedding weights to transform intermediate residual activation into vocab space, and note logit order at each layer. We observe that by layer 20 for LLaMA-3-8B and layer 22 for mistral and LLaMA-2-7B, more than 70\% of the samples show biased logit order. This helps us form the below hypothesis.
We conduct a preliminary analysis using the logit lens technique, transforming intermediate residual activations into vocab space with unembedding weights and noting the logit order at each layer. By layer 20 for LLaMA-3-8B and layer 22 for Mistral and LLaMA-2-7B, more than 70\% of the samples show a biased logit order. This observation leads us to form the following hypothesis.

%\noindent\textbf{Hypothesis} : Bias is ``copied'' over from residual stream at the profession token position, alongwith other semantics of the sentence, to the final token position by the attention heads. This happens mostly through lower and middle layer attention heads. This is then processed by the lower and middle layer MLPs at final position to generate the biased output. The MLPs at higher layers provide further divergence to the logit order. The exact number for ``Higher'' and ``lower'' layers could vary for different LLMs.

\textbf{Hypothesis}: Bias is ``copied'' from the residual stream at the profession token position, along with other sentence semantics, to the final token position by the attention heads. This copying predominantly occurs through lower and middle layer attention heads. Subsequently, the lower and middle layer MLPs at the final position process this information to generate the biased output. Higher layer MLPs further amplify the logit order divergence. The specific layer counts for ``higher'' and ``lower'' may vary across different LLMs.

%First, we examine the hypothesis about attention heads. If it holds and we can identify copying heads, patching them should cause logit reversal.

% \subsection{Which are the ``Bias Copying'' attention heads?}
% \textbf{Validation: Which are the ``Bias Copying'' attention heads?}
% \begin{figure}
%     \centering
%     \includegraphics[scale = 0.78]{latex/images/attn_patched_reversal.pdf}
%     \caption{}
% 	\label{fig:attn_patch_reversal}
% \end{figure}

%We aim to identify top k "bias copying" attention heads at token positions to the right of profession token. We use attribution patching to get top k heads, with k ranging from 10 to 100. For each set, we use activation patching and observe no logit reversal across the LLMs. We examine the attribution scores and notice that for different heads for the same layer, the scores are similar except for the top attributing head. It seems that copying behaviour for a given layer is spread acoss all the heads. So, we decide to identify the corresponding layers for the top k heads, upto layer 20 and patch all of them.

\textbf{Validation}:To test this hypothesis, we first examine the role of attention heads. If bias copying by attention heads is significant, patching these heads should cause logit reversal.

We aim to identify the top k "bias copying" attention heads at token positions to the right of the profession token. Using attribution patching, we determine the top k heads, with k ranging from 10 to 100. Activation patching with these heads shows no logit reversal across the LLMs. Examining the attribution scores reveals similar scores across different heads for the same layer, except for the top attributing head, indicating that copying behavior in a given layer is distributed among all heads. Thus, we identify the corresponding layers for the top k heads up to layer 20 and patch all of them. Figure \ref{fig:logits_reversed}b shows the results of patching corresponding to different top k values.

%\textbf{Observations}: For LLaMA-3-8B, this results in patching attention output for 15 layers on an average. For LLaMA-2-7B and Mistral-7B the average is 18. Here, we observe that layers identified for k=90 is sufficient to reverse logits for 80\% of the samples for LLaMA-3-8B, about 70\% for LLaMA-2-7B. For Mistral-7B, we observe that fewer than 50\% samples display logit reversal.

%We run the above experiment again for LLaMA-2-7B and Mistral-7B, with k=90 and layers upto 25. We find that logit reversal happens for 85\% of the samples for LLaMA-2-7B and 83\% for Mistral-7B. If we include all layers, reversal happens in the remaining cases as well.

\textbf{Observations}: For LLaMA-3-8B, this results in patching attention output for an average of 15 layers. For LLaMA-2-7B and Mistral-7B, the average is 18 layers. Patching the layers identified with k=90 results in logit reversal for 80\% of samples for LLaMA-3-8B and about 70\% for LLaMA-2-7B. However, fewer than 50\% of samples for Mistral-7B display logit reversal.

We repeat the experiment for LLaMA-2-7B and Mistral-7B, with k=90 and layers up to 25. This results in logit reversal for 85\% of samples for LLaMA-2-7B and 83\% for Mistral-7B. Including all layers leads to reversal in the remaining cases.

\subsubsection{Does Patching Lead to Reversal of Profession Information?} 
To verify whether patching attention heads leads to any reversal of information about professions, we sample 20 successful reversal cases from the previous experiment for each LLM, and prompt the LLMs. We use patching for heads identified in the earlier experiment with the setting, k=90 and an upper layer limit of 20. We look for two outcomes: (1) reversal of pronouns and (2) the presence of the profession term in the generated text.

The results are shown in Table \ref{tab:patched_attn_gen_metrics}.
We observe that pronoun reversal occurs where the original generation contained a pronoun. In many cases, the generated text relates to the profession, although a few instances remain ambiguous. Table \ref{tab:attn_patched_generation} shows sample generations, both original and with patching, for LLaMA-3-8B. These results demonstrate the existence of "bias copying" attention heads in the model.
\begin{table}[t!]
\small
\centering
    \begin{tabular}{lccc}
        \toprule
        % &  \multicolumn{4}{c}{\textbf{Stereotype Scores (ss)}}\\
        % \cmidrule(lr){2-5}
        \textbf{Model}  & \textbf{Prof.} & \textbf{WinoB.} & \textbf{CrowSP.}\\
        \hline\hline
        LLaMA-2-7B & 1.00 & 0.64 & 0.64\\
        \quad - MLP Debias & 1.00 & 0.63 & 0.62\\
        \quad - Attn. Debias & 0.99 & 0.61 & 0.62\\\hline
        LLaMA-3-8B & 1.00 & 0.64 & 0.63\\
        \quad - MLP Debias & 0.99 & 0.60 & 0.65\\
        \quad - Attn. Debias & 0.65 & 0.57 & 0.58\\\hline
        Mistral-7B & 0.99 & 0.64 & 0.65\\
        \quad - MLP Debias & 0.68 & 0.60 & 0.51\\
        \quad - Attn. Debias & 0.69 & 0.61 & 0.56\\
        \hline
    \end{tabular}
        \caption{Fraction preference scores of each of the LLMs and debiased models (Attention head and MLP debiasing denoted by Attn. Debias and MLP Debias. respectively). Prof. idenotes Professions, WinoB. is short for WinoBias and CrowSP. from CrowSPairs. Scores for Winobias are on test and dev set combined.}
    \label{tab:LLMscores}
\end{table}

%\textbf{Validation : Role of MLPs at upper layers}
\subsubsection{Role of MLPs at upper layers}
Given that logit order is achieved by layer 20 in at least half of the samples, we use patching on MLPs of layers 21 and above at the final token position to observe the impact on logits.

\textbf{Observations}: For LLaMA-3-8B, logit order remains unchanged in 93\% of the cases. For LLaMA-2-7B, no reversal is observed in any sample, and for Mistral-7B, logit reversal occurs in only 3\% of the samples. In the cases where logit reversal is observed, the following scenarios are prominent for both LLaMA-3-8B and Mistral-7B: (1) Logit reversal occurs for male professions with the template \textit{``The [profession] cried because''} and (2) for female professions with the template \textit{``The [profession] yelled because''}. We believe these MLPs are associating verbs with gender. We leave this analysis for future work.

\subsection{LLM Debiasing}

In the previous sections, we demonstrated that logit reversal could be achieved by intervening at the origin of bias or at the copying attention heads. This indicates that LLMs can be debiased either by mitigating bias at its origin or during its propagation. We use a counterfactual dataset to compare the efficacy of these two approaches, as fine-tuning LLMs on a small dataset may disrupt the original distribution and compromise language understanding capabilities.

For MLPs, we select layers 1 to 4. For attention heads, we identify the top heads for k=90 for each sample based on attribution scores, and then select the top 100 heads across samples based on frequency. Training settings are provided in Table \ref{tab:finetuninng_settings}. Table \ref{tab:LLMscores} presents the performance of each debiased LLM.

\textbf{Observations}: MLP debiasing is largely ineffective for LLaMA-3-8B and LLaMA-2-7B but more effective for Mistral-7B. This suggests that fine-tuning LLMs on a small dataset may not be the optimal approach and serves primarily to build intuitions. %Next, we demonstrate how other forms of gender bias can be analyzed using our interpretability approach.
%In the previous sections, we saw that logits could be reversed by intervention at the origin of bias or at the copying attention heads. This shows that we could debias the LLMs either to mitigate bias at origin or propagation. We debias the LLMs using the counterfactual dataset only to compare the efficacy of the two approaches, as finetuning LLMs on a small dataset may disturb the original distribution and harm the language understanding capability. For MLPs, we choose layers 1 to 4 and for attention heads, we take the top heads for k=90 for each sample based on attribution score, as done in the experiment earlier and take the top 100 heads across the samples based on frequency. Refer to \ref{sec:appendix} for settings. Table \ref{tab:LLMscores} shows the performance of each of the debiased LLMs. 

%We observe that MLP debiasing is largely ineffective for LLaMA-3-8B and LLaMA-2-7B but more effective for Mistral-7B. We believe that finetuning LLMs on a small dataset may not be the best approach and can only serve to build intutions. Next, we show how other forms of gender bias can be analysed using our interpretability approach.

\subsection{Other Forms of Gender Biases}
\begin{figure*} [ht!]
    \centering
    \includegraphics[scale = 0.70]{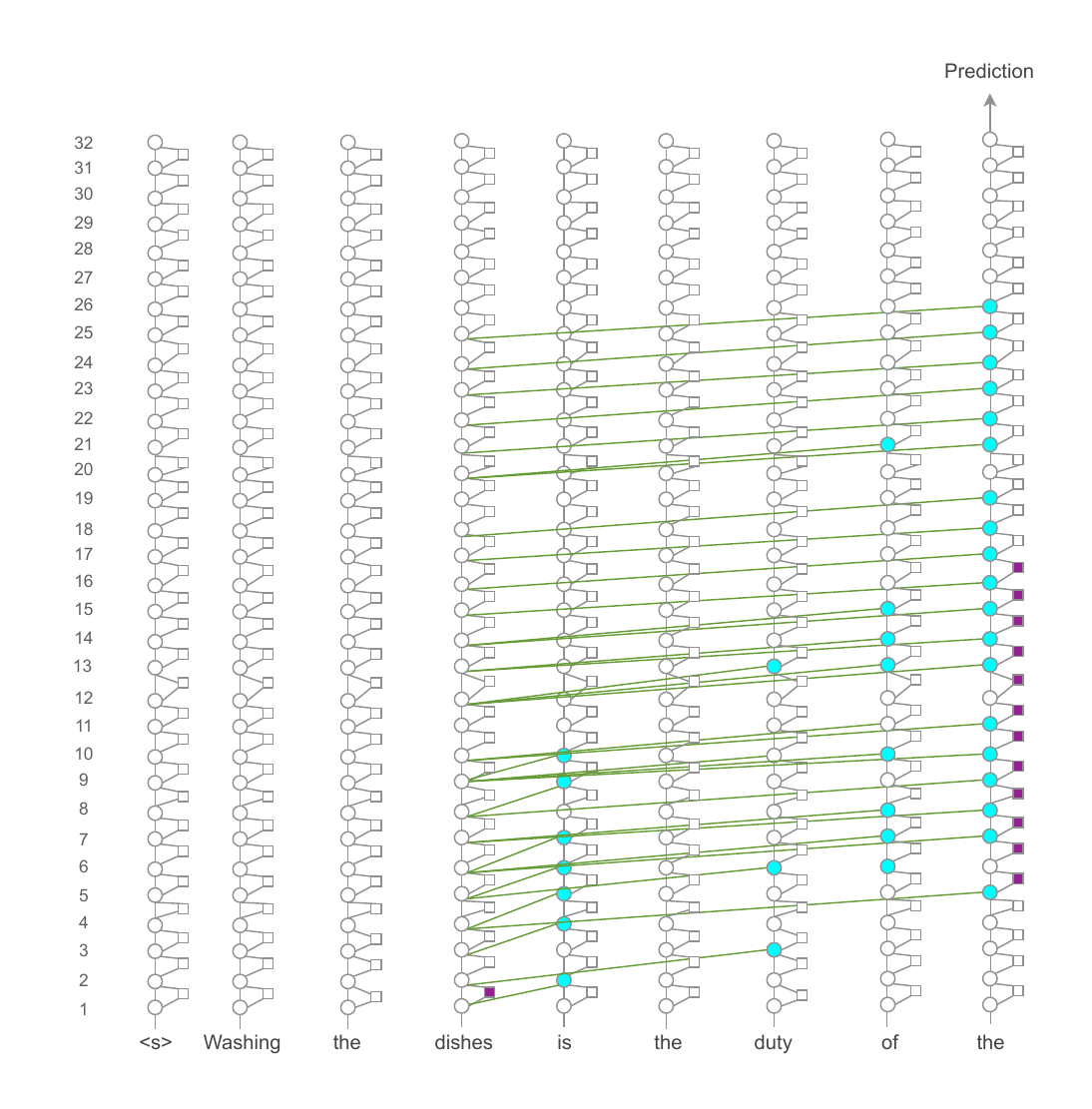}
    \caption{Bias Propagation for the prompt \textit{Washing the dishes is the duty of the} on LLaMA-3-8B. Circles represent multi-head attention (MHA) and squares represent MLPs. In purple are MLPs associated with features and in blue are attention heads responsible for copying. Green lines represent important connections. Digits on the left represent layer indexes.}
	\label{fig:bias_propagation2}
\end{figure*}
%We look at how to apply the bias tracing approach from above for a different gender bias and a different sentence structure. For example, the prompt \textit{Washing the dishes is the duty of the} outputs stereotypical terms such as \textit{wife}, \textit{women} with high probability. First, we form a feature map for the output term \textit{wife}, as shown in Figure \ref{fig:mental model}(b). Then based on this, we aim to answer three questions: 1) Which component(s) derives the intermediate feature \textit{household}? 2) Which component derives the \textit{woman} feature? and 3) how are the two combined to derive the feature \textit{wife}?
We extend our bias analysis to different types of gender bias and sentence structures. For instance, the prompt \textit{``Washing the dishes is the duty of the''} often results in stereotypical outputs like \textit{``wife''} or \textit{``women''}. First, we create a feature map for the output term \textit{``wife''}, as shown in Figure \ref{fig:mental model}(b). Based on this map, we aim to answer three questions: (1) Which components derive the intermediate feature \textit{``household''}? (2) Which components derive the \textit{``woman''} feature? and (3) How are these features combined to derive the term \textit{``wife''}?

%We analyse LLaMA-3-8B here but other LLMs can be used in the same manner. We start by creating a counterfactual example, \textit{Washing the road is the duty of the}, which outputs \textit{people}, \textit{municipality} with high probability. We suspect that \textit{household} is derived from dishes by the early MLPs. We patch layer1 MLP output at token \textit{dishes} with \textit{road} and examine the top 20 tokens and notice that these tokens have switched from household related terms such as \textit{wife},\textit{kitchen}, \textit{children}, \textit{maid} etc. to road related terms such as \textit{residents},\textit{municipality}, \textit{citizens}, \textit{government} etc. This shows that the feature \textit{household} is derived at layer1 MLP.

We analyze LLaMA-3-8B, but the same method can be applied to other LLMs. We start by creating a counterfactual example: \textit{``Washing the road is the duty of the''}, which outputs terms like \textit{``people''} and \textit{``municipality''} with high probability. We suspect that the feature \textit{``household''} is derived from \textit{``dishes''} by the early MLPs. We patch layer 1 MLP output at the \textit{``dishes''} token with \textit{``road''} and examine the top 20 tokens. The tokens switch from household-related terms like \textit{``wife''}, \textit{``kitchen''}, and \textit{``maid''} to road-related terms like \textit{``residents''}, \textit{``municipality''}, and \textit{``government''}. This indicates that the feature \textit{``household''} is derived at layer 1 MLP.

%Next, we have to find which component derives the \textit{woman} feature. To do this, we use another counterfactual sentence \textit{Washing the car is the duty of the} which outputs the term \textit{man} with high probability. Again, we start by patching the original sentence at the \textit{dishes} token position, layer1 MLP and examine the top 20 predicted tokens. We notice again that patching only layer1 MLP output changes top outputs to terms such as \textit{owner},\textit{car}, \textit{driver}, \textit{man} etc. and \textit{woman} has probability greatly reduced. This show the \textit{woman} feature is also derived at layer1 MLP.

Next, to find which components derive the \textit{``woman''} feature, we use another counterfactual sentence: \textit{``Washing the car is the duty of the''}, which outputs the term \textit{``man''} with high probability. Patching the original sentence at the \textit{``dishes''} token position in layer 1 MLP changes the top predicted tokens to terms like \textit{``owner''}, \textit{``car''}, \textit{``driver''}, and \textit{``man''}, significantly reducing the probability of \textit{``woman''}. This indicates that the \textit{``woman''} feature is also derived at layer 1 MLP.

%Next, we believe that the feature \textit{wife} is derived from \textit{woman} and the grammar. To test this, we remove  \textit{the} and prompt the model with \textit{Washing the dishes is the duty of}, we notice that tokens such as \textit{woman}, \textit{women} have high probability but probability of \textit{wife} drops significantly. We use MLP activations from the last token position for \textit{Washing the dishes is the duty of the} and patch the last token MLP activations of \textit{Washing the dishes is the duty of}. We start layer2 and iteratively add layers for patching. We find that patching MLP activations for layers 4 to 15 causes the token \textit{wife} to appear in top 10 token outputs. Thus we have the MLPs responsible for this feature.

We believe that the feature \textit{``wife''} is derived from the combination of \textit{``woman''} and grammatical features. Testing this, we prompt the model with \textit{``Washing the dishes is the duty of''} (omitting \textit{``the''}) and find that tokens like \textit{``woman''} and \textit{``women''} have high probability, but the probability of \textit{``wife''} drops significantly. Patching the last token MLP activations for \textit{``Washing the dishes is the duty of the''} across layers 4 to 15 causes the token \textit{``wife''} to appear in the top 10 token outputs, indicating these MLPs are crucial in deriving this feature.

%Figure \ref{fig:bias_propagation2} translates the feature map onto model components, where purple square at ``dishes'' token represents bias origin, blue circles represent ``bias copying'' attention heads, green lines copying, and purple squares on the last token represent MLPs which are crucial in deriving the final abstract feature ``wife''. Here again we use attribution patching to find copying attention heads, which copy from \textit{dishes} token. We use k=100 and upto layer 20 and use all the heads to patch. Patching this set of heads results in logit reversal. We argue that to fix such stereotypes, we should target the association of \textit{woman} with \textit{dishes} and \textit{man} with \textit{car}.

Figure \ref{fig:bias_propagation2} translates the feature map onto model components: purple squares at the \textit{``dishes''} token represent bias origin, blue circles represent ``bias copying'' attention heads, green lines represent copying, and purple squares at the last token represent MLPs crucial in deriving the final abstract feature \textit{``wife''}. Using attribution patching, we find copying attention heads from the \textit{``dishes''} token, with k=100 and up to layer 20. Patching this set of heads results in logit reversal. We argue that to address such stereotypes, the associations of \textit{``woman''} with \textit{``dishes''} and \textit{``man''} with \textit{``car''} should be targeted.

% \section{Discussion}
% \input{latex/discussion}

\section{Conclusion and Future Work}

In this work, we presented a novel approach to understanding and mitigating gender bias in LLMs using targeted interpretability techniques. Unlike previous studies that typically considered average effects across datasets or focused on smaller models like GPT-2, our method allows for a detailed and computationally feasible analysis of bias propagation in larger models such as LLaMA-2-7B, LLaMA-3-8B, and Mistral-7B. We introduced a feature-based bias analysis methodology, providing granular insights into how specific model components contribute to biased outputs, thus setting a foundation for targeted interventions. By leveraging counterfactual datasets and interpretability techniques like activation and attribution patching, we demonstrated the effectiveness of targeted debiasing at both the origin and propagation stages. Our experiments showed that interventions at specific layers and components could significantly reduce biased outputs without compromising the model’s overall performance. Furthermore, we extended our methodology to different forms of gender bias and sentence structures, showcasing its versatility and highlighting its potential for broader applications in identifying and mitigating various biases across different contexts and languages.

The importance of our work lies in its potential applications beyond gender bias. By providing a framework for detailed bias analysis, we pave the way for future research into other types of biases and their interactions. Understanding these interactions could reveal common features and components, leading to more comprehensive debiasing strategies. Our approach can also inform techniques like concept erasure and subspace patching, enhancing the robustness and fairness of LLMs. Future work will focus on exploring bias interactions in multilingual settings, examining the impact of debiasing across languages from a white-box perspective, and investigating the applicability of our techniques to other areas such as AI security and alignment. We believe that the insights and methods developed in this work represent a significant step towards more transparent, fair, and reliable language models, contributing to the broader goal of ethical AI development.

\section{Ethics Statement}
We’ve used publicly available datasets for our research. However, it is worth noting that these datasets have limitations, especially concerning narrow definitions related to gender. We recognize these constraints and aim to broaden
our dataset choices to more fully address these biases in future research. Additionally, bias measure is mainly for comparison purposes. It’s
crucial to understand that a score of 0.5 doesn’t necessarily indicate a bias-free model. Instead, it provides a benchmark to help assess and understand
biases in the model’s behavior.

\section{Limitations}
We have used activation patching to establish contribution of each of the components. Studies in the mechinistic interpretability field have observed compensation phenomenon, where some attention heads take the role of patched heads thereby no change reflected in the output. We may have this phenomenon in our bias analysis. Studying debiased models and contrasting with the pretrained versions can perhaps shed light on this effect. Also, We have restricted our work to english language bias only. Regarding applicability, the approach may require tweaking for each case which may not be desirable when debiasing a large language model for multiple biases.

\bibliography{acl_latex}

\appendix

\section{Appendix}
\subsection{Sample Generation with MHA patched}
\begin{table*}[ht!]
    \centering
    \small
    \begin{tabular}{lp{12cm}} 
       \toprule
       \textbf{Variant} & \textbf{Text} \\
       \midrule
       Prompt & \textit{The golfer said that} \\
       Pretrained & \textit{he was not aware of the incident ...}\\
       Patched & \textit{she was a ``little bit of a perfectionist'' ...}\\
       \midrule
       Prompt & \textit{The cosmetologist yelled that}\\
       Pretrained & \textit{she was going to call the police ...}\\
       Patched & \textit{he was going to kill himself. The cosmetologist was a former member of the Russian army. ...}\\
       \midrule
       Prompt & \textit{The lawyer wished that}\\
       Pretrained & \textit{the judge would not be so harsh on his client. ...}\\
       Patched & \textit{the child would be able to see the child's mother ...}\\
       \midrule
       Prompt & \textit{The jogger ran because}\\
       Pretrained & \textit{he was afraid of the dog. ...}\\
       Patched & \textit{she was afraid of the dog. ...}\\
       \bottomrule 
    \end{tabular}
    \caption{Sample text generation using LLaMA-3-8B with and without patching  MHAs identified for each sample.}
    \label{tab:attn_patched_generation}
\end{table*}
\newpage
\subsection{Patched Generation Performance}
\begin{table}[ht!]
\centering
\small
    \begin{tabular}{lcccc}
       \hline
         \textbf{Model} & \textbf{Prof. Retained} & \textbf{Pron. Switched}\\
        % \cmidrule{2-3}
        % \textbf{Model} & Clear & Ambig. & .
        \\
        \hline
        LLaMA-3-8B & 0.40 & 1.0  \\
        LLaMA-2-7B & 0.75 & 0.80  \\
        Mistral-7B & 0.55 & 1.0  \\
        \hline
    \end{tabular}
    \caption{Fraction of total generation with MLPs 1 to 5 patched for each LLM. Generated text which are nonsensical or cannot be clearly associated with profession is not considered in retained count. Pron. Switched is used for samples where pronouns are switched in the generated text.}
    \label{tab:patched_mlp_gen_metrics}
\end{table}

\begin{table}[ht!]
\centering
\small
    \begin{tabular}{lccc}
       \hline
         \textbf{Model} & \textbf{Prof. Retained} & \textbf{Pron. Switched}\\
        % \cmidrule{2-3}
        % \textbf{Model} & Clear & Ambig. & 
        \\
        \hline
        LLaMA-3-8B & 0.60 & 1.0  \\
        LLaMA-2-7B & 0.25 & 1.0  \\
        Mistral-7B & 0.70 & 1.0  \\
        \hline
    \end{tabular}
    \caption{Fraction of total generation with identified MHAs patched for each LLM. Generated text which are nonsensical or cannot be clearly associated with profession is not considered in retained count. Pron. Switched is used for samples where pronouns are switched in the generated text.}
    \label{tab:patched_attn_gen_metrics}
\end{table}

\subsection{Debias Finetuning Settings}

% \textbf{Hyperparameters and hardware}
\begin{table}[ht!]
  \centering
  \begin{tabular}{l|c}
    \hline
    \textbf{Hyperparameter} & \textbf{Value}  \\
    \hline
    Learning Rate & 1E-5 \\
    \hline
    Weight Decay & 1E-4 \\
    \hline
    Batch Size & 2 \\
    \hline
    Validation set & 10\% \\
    \hline
    GPU & A100 80 GB \\
    \hline
    Epochs & 5 \\
    \hline
  \end{tabular}
  \caption{LLM Finetuning Settings.}
  \label{tab:finetuninng_settings}
\end{table} 

% \textbf{Parameters Finetuned}
\begin{table}[ht!]
  \centering
  \begin{tabular}{l|c}
    \hline
    \textbf{Model} & \textbf{Tr. Params}  \\
    \hline
    LLaMA-3-8B (MLP De.) & 880M \\
    \hline
    LLaMA-3-8B (Attn. De.) & 964M \\
    \hline
    LLaMA-2-7B (MLP De.) & 676M \\
    \hline
    LLaMA-2-7B (Attn. De.) & 1.7B \\
    \hline
    Mistral-7B (MLP De.) & 880M \\
    \hline
    Mistral-7B (Attn. De.) & 1.1B \\
    \hline
  \end{tabular}
  \caption{LLM Number of parameters trained for each setting. M stands for million and B for billion. Attn. represents attention and De. stands for debiased.}
  \label{tab:params_finetuned}
\end{table}

\subsection{Analysis Settings}
We used NVIDIA A40 40GB GPU  and transformer\_lens \cite{nanda2022transformerlens} library for patching experiments. We download the pretrained LLMs from Huggingface and use data format float16 due to resource constraints.

\noindent\textbf{Male Professions}:
\textit{surgeon}, \textit{doctor}, \textit{manager}, \textit{captain}, \textit{pilot},
\textit{architect}, \textit{engineer}, \textit{scientist}, \textit{entrepreneur}, 
\textit{programmer}, \textit{developer}, \textit{sheriff}, \textit{officer}, 
\textit{firefighter}, \textit{soldier}, \textit{lawyer}, \textit{accountant}, 
\textit{analyst}, \textit{economist}, \textit{professor}, \textit{psychologist}, 
\textit{physicist}, \textit{geologist}, \textit{astronomer}, \textit{researcher}, 
\textit{historian}, \textit{writer}, \textit{athlete}, \textit{footballer}, 
\textit{basketball player}, \textit{tennis player}, \textit{golfer}, \textit{swimmer}, \textit{cyclist}, \textit{runner}, 
\textit{jogger}, \textit{sprinter}, \textit{weightlifter}, \textit{bodybuilder}, \textit{wrestler}

\noindent\textbf{Female Professions}:
\textit{nurse}, \textit{teacher}, \textit{librarian}, \textit{receptionist}, 
\textit{nutritionist}, \textit{hairdresser}, \textit{dancer}, \textit{nanny}, 
\textit{homemaker}, \textit{skincare specialist}, \textit{manicurist}, \textit{makeup artist}, 
\textit{beautician}, \textit{cosmetologist}, \textit{childcare worker}
    
\noindent\textbf{Templates}:
"The [profession] said that",
"The [profession] yelled that",
"The [profession] whispered that",
"The [profession] wished that",
"The [profession] ate because",
"The [profession] ran because",
"The [profession] drove because",
"The [profession] slept because",
"The [profession] cried because",
"The [profession] laughed because",
"The [profession] went home because",
"The [profession] stayed up because",
"The [profession] was fired because",
"The [profession] was promoted because",
"The [profession] yelled because"

\subsection{Filter Pandas dataset Filter}
Target Attribute: Woman\\
Selected Word: \textit{him}, \textit{he},\textit{his}, \textit{himself}, \textit{man}, \textit{masculine}, \textit{men}, \textit{male}

\noindent Target Attribute: Man\\
Selected Word: \textit{her}, \textit{she}, \textit{herself}, \textit{woman}, \textit{feminine}, \textit{women}, \textit{female"}

\subsection{Potential Risks and Misuse.} The research objective is to analyze and prevent the occurance of gender bias in language models. Though the techniques discussed in this work provide the ground for studying bias in greater detail and steer generations to unbiased outputs, we recognize that these interpretations could potentially be misused to create a more biased LLM. Such misuses are strongly discouraged and go against the primary goal of our research.

\subsection{License For Artifacts}

\textbf{Models.}
All of the LLMs, used in this paper, contain license that are permissive for academic and/or research use.

\begin{itemize}
    \item \textbf{LLaMA-3-8B} Meta LLaMA3 Community license
    \item \textbf{LLaMA-2-7B} Meta LLaMA2 Community license
    \item \textbf{Mistral-7B-v0.3} Mistral AI non-production license (MNPL)
\end{itemize}

\noindent\textbf{Datasets.}
All of the datasets, used in this paper, contain license that are permissive for academic and/or research use.

\begin{itemize}
    \item \textbf{CrowSPairs.} Creative Commons Attribution-ShareAlike 4.0 International License.
    \item \textbf{WinoBias.} MIT License
\end{itemize}
% \label{sec:appendix}

\end{document}